\documentclass{svproc}

\usepackage{hyperref}
\usepackage{url}

\usepackage{booktabs}
\usepackage{graphicx}
\usepackage{amsmath}
\usepackage{amssymb}
\usepackage{pgfplots}
\pgfplotsset{compat=1.14}
\usepackage{adjustbox}
\usepackage{array}
\usepackage{makecell}
\usepackage{siunitx}
\usepackage{breakcites}
\usepackage{enumitem}
\usepackage{comment}
\usepackage{makecell}

\newcommand{\usd}[1]{USD~#1}

\renewcommand{\subsubsection}[1]{\textbf{#1}}

\newcolumntype{R}[2]{%
    >{\adjustbox{angle=#1,lap=\width-(#2)}\bgroup}%
    l%
    <{\egroup}%
}

\begin{document}
\mainmatter
\title{The Cambridge RoboMaster:\texorpdfstring{\newline}{} An Agile Multi-Robot Research Platform}
%
\titlerunning{Cambridge RoboMaster for Multi-Robot Research}  
%
%
\author{Jan Blumenkamp \and Ajay Shankar \and \\
Matteo Bettini \and Joshua Bird \and Amanda Prorok}
\authorrunning{Blumenkamp et al.} 
%
\tocauthor{Jan Blumenkamp, Ajay Shankar, Matteo Bettini, Joshua Bird, Amanda Prorok}
\institute{Department of Computer Science and Technology\\ University of Cambridge\\
United Kingdom\\
\email{\{jb2270, as3233, mb2389, jyjb2, asp45\}@cst.cam.ac.uk}}

\maketitle              

\begin{abstract}

Compact robotic platforms with powerful compute and actuation capabilities are key enablers for practical, real-world deployments of multi-agent research.
This article introduces a tightly integrated hardware, control, and simulation software stack on a fleet of holonomic ground robot platforms designed with this motivation.
Our robots, a fleet of customised DJI Robomaster S1 vehicles, offer a balance between small robots that do not possess sufficient compute or actuation capabilities and larger robots that are unsuitable for indoor multi-robot tests.
They run a modular ROS2-based optimal estimation and control stack for full onboard autonomy,
contain ad-hoc peer-to-peer communication infrastructure,
and can zero-shot run multi-agent reinforcement learning (MARL) policies trained in our vectorized multi-agent simulation framework.
We present an in-depth review of other 
platforms currently available, showcase
new experimental validation of our 
system’s capabilities, and introduce
case studies that highlight the
versatility and reliability of our system as a 
testbed for a wide range of research
demonstrations.
Our system as well as supplementary material is available online.
\footnote{\url{https://proroklab.github.io/cambridge-robomaster}}

\end{abstract}

\section{Introduction}

Multi-agent robotics research necessitates robotic platforms that can be used as deployment testbeds for rapid development and evaluations~\cite{hyldmar_fleet_2019,caprari_design_2003,wang_shape_2020,arvin_mona_2019,mondada_e-puck_2009,arvin_development_2009,kernbach_re-embodiment_2009,mclurkin_robot_2014,wilson_pheeno_2016,paull_duckietown_2017,soares_khepera_2016,giernacki_crazyflie_2017,ozgur_cellulo_2017,arvin_development_2014,blumenkamp_framework_2022,bettini_heterogeneous_2023,blumenkamp_see_2023,gao_online_2023,yang_efficient_2023}.
The choices for such platforms are usually informed by several factors such as the configuration space of the problem, the number of agents involved, and the computational sophistication expected from each agent.
For instance, small-scale robots, such as the Turtlebot~\cite{turtlebot3_turtlebot3_2024}, will frequently be limited in their agility of motion and are restricted in their onboard computing capabilities.
On the other hand, while larger robots (Jackal~\cite{clearpath_robotics_inc_jackal_2020} or AutoRally~\cite{williams_aggressive_2016}) have greater manoeuvrability and can carry more compute payloads, operating multiple of them in smaller research spaces can be cost- and space- prohibitive.
Successfully testing and developing multi-robot solutions requires platforms that strike a fine balance between these objectives.

\begin{figure}[t]
    \centering
    \includegraphics[height=2cm]{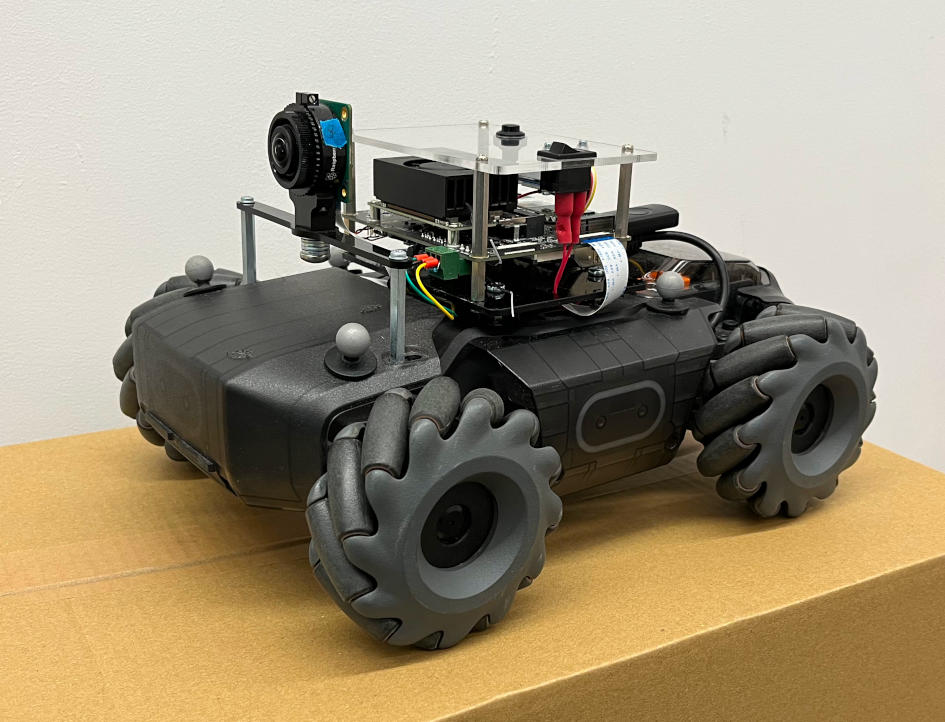}
    \hfill
    \includegraphics[height=2cm]{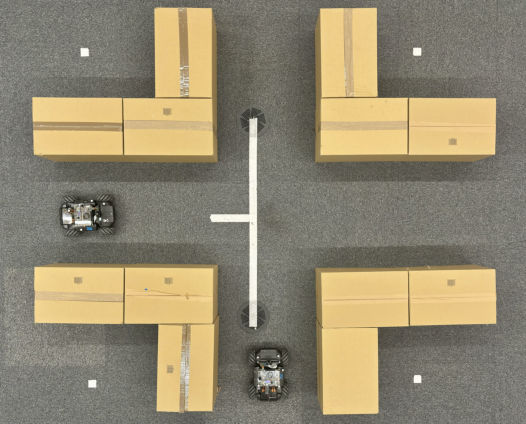}
    \hfill
    \includegraphics[height=2cm]{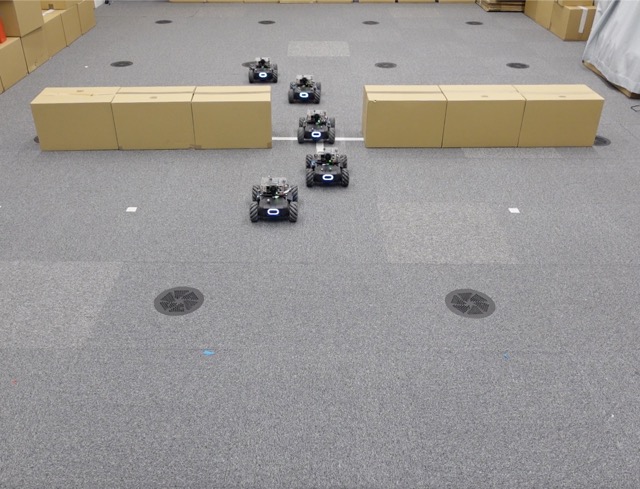}
    \hfill
    \includegraphics[height=2cm]{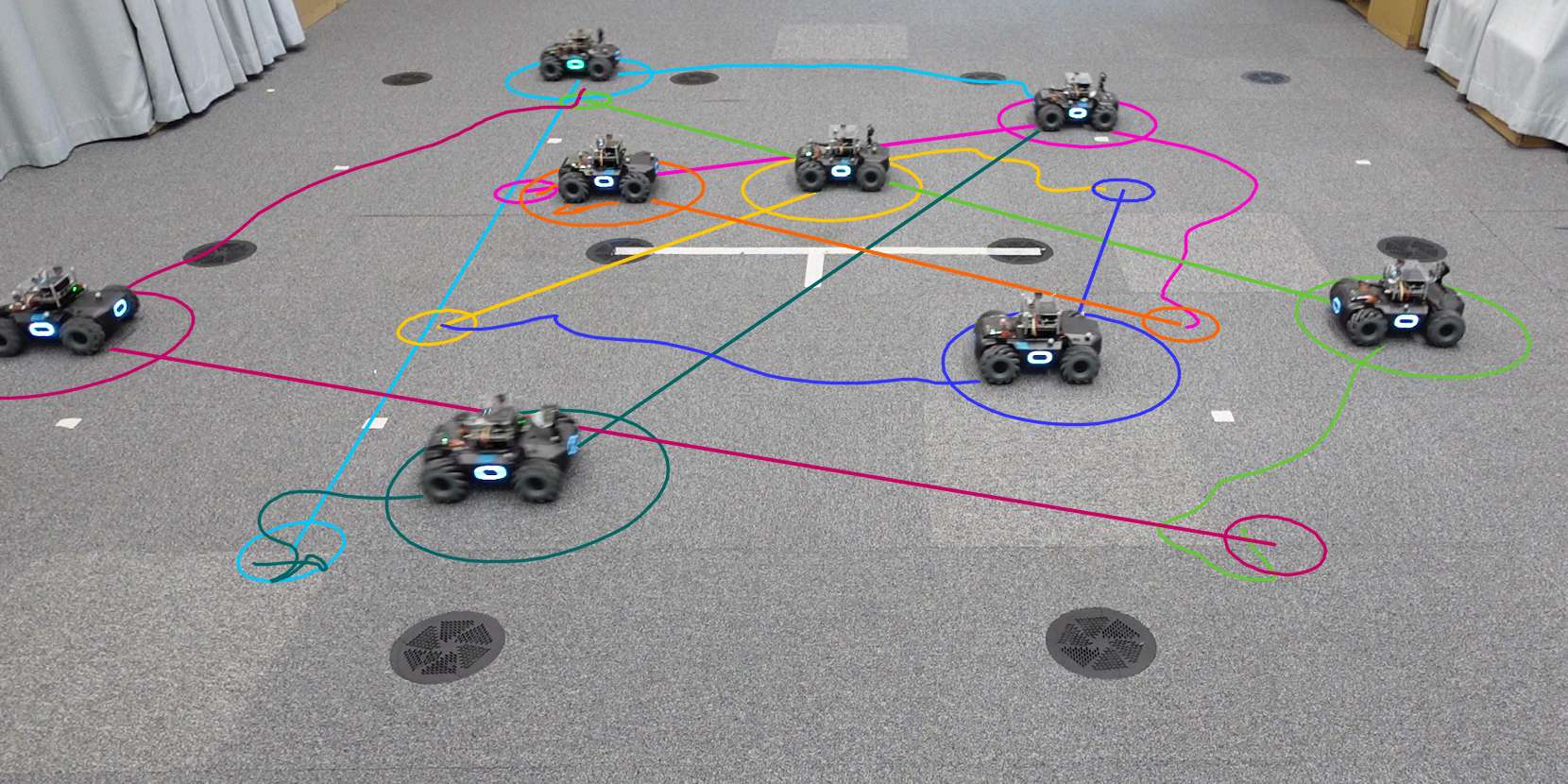}
    \caption{Shown from left to right: 1) A close-up of one RoboMaster equipped with an NVidia Jetson Orin NX and a forward-facing camera 2) Two RoboMasters in a SLAM-based collision avoidance scenario 3) Five RoboMasters moving through a narrow constriction after breaking formation~\cite{blumenkamp_framework_2022} 4) Eight RoboMasters in a multi-robot navigation scenario trained in VMAS~\cite{bettini_vmas_2022}.}
\label{fig:hero}
\end{figure}

In this article, we present an agile and affordable system and platform, shown in \autoref{fig:hero} as a multi-robot team setup.
Our system, an omnidirectional indoor ground robot based on the DJI Robomaster S1, is
\textit{affordable} (starting at \usd{689} in the base configuration to \usd{1633} with all sensors and an Jetson Orin NX),
\textit{agile} (we demonstrate indoor trajectory tracking capabilities of up to \SI{4.45}{m/s} and accelerations of up to \SI{5}{m/s^2}),
\textit{versatile} (we show experiments from classical control to zero-shot deployment of MARL policies trained in our simulator VMAS~\cite{bettini_vmas_2022}),
and is easily \textit{replicable} (our entire stack is open-sourced, requiring minimal construction time).
These have been made possible due to our retrofitting of the onboard compute module and careful reverse engineering of the main communication interface.
We present two further contributions in this work. First, through an expansive review of related systems and literature, we contrast our solution against others' and position our system at the frontier of low-cost agile research platforms.
Second, we describe enhancements to our prior work (VMAS~\cite{bettini_vmas_2022}) that enable rapid deployment of control policies learnt in simulation onto this platform.
We present evaluations and case studies that corroborate the claims presented above, and showcase the wide range of research avenues that this system makes accessible.

\section{Related Work}
\begin{figure}[t]
    \centering
    \newcommand{\mathdefault}[1][]{}
    \input{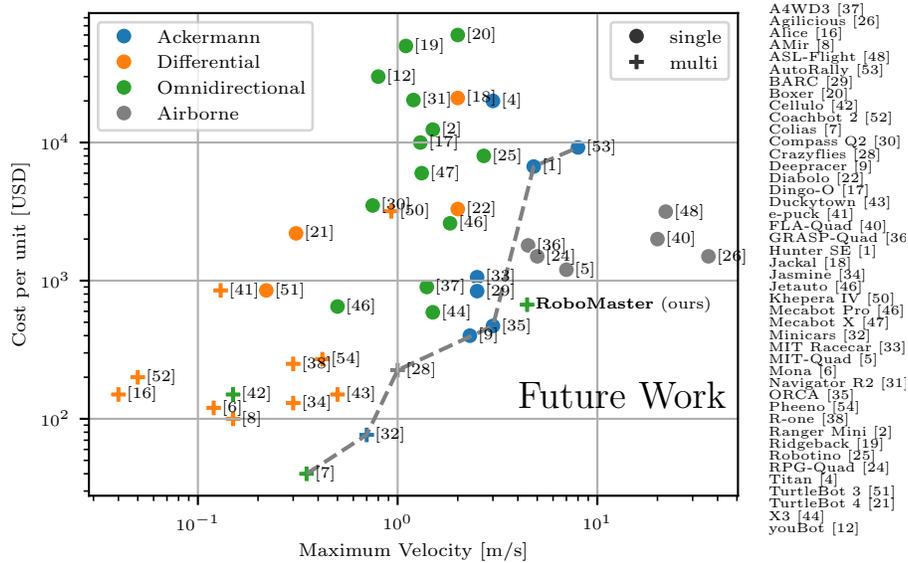}
    \caption{We compare a variety of different research and commercial robotic platforms concerning their tradeoff between cost per unit and maximum velocity. We differentiate between Ackermann, Differential, Airborne and Omnidirectional platforms as well as between platforms specifically targetting multi-agent research. A clear correlation can be seen between speed and cost, as well as a trend for multi-agent platforms to be low-cost, but as a consequence slow. We draw the Pareto front as a dashed line, with our proposed platform pushing the limits on the cost and speed tradeoff.}
    \label{fig:robot_platforms}
\end{figure}

In the last decade, a wide range of (multi-) robot platforms have been introduced to the research community.
Here we present an in-depth literature review of commercial as well as research platforms.
Our classification scheme will differentiate between them based on their dynamics (Ackermann, differential, and omnidirectional drive, as well as airborne/multirotor platforms), and also based on their specific use cases in single- or multi-agent research.
Since we are primarily interested in multi-robot research, we investigate the tradeoff between cost and speed of these platforms, as agility is an attribute we
require in the team (note that we use `speed' and `agility' interchangeably -- this is a justified simplification for indoor robots operating in confined spaces), and cost limits the size of the team.

\begin{table}[t]
    \scriptsize
    \centering
    \setlength{\tabcolsep}{3pt}
    \caption{We list critical specifications for omnidirectional robots from related work. `Multi-Agent' indicates whether the platform is introduced as a multi-agent platform, and `Commercial' indicates whether the platform can be obtained commercially.}
    \label{tab:omnidir_robots}
    \begin{tabular}{llrrrccc}
\toprule
Brand/Uni & Product & \makecell{Vel \\ {[m/s]}} & \makecell{Cost \\ {[USD]}} & \makecell{Mass \\ {[kg]}} & \makecell{L x B x H\\ {[mm]}} & \makecell{Multi \\ {Agent}} & Commercial \\
\midrule
EPFL & Cellulo \cite{ozgur_cellulo_2017} & 0.15 & 150.0 & 0.18 & $73 \times 80 \times 80$ & \checkmark &  \\
Lincoln Uni & Colias \cite{arvin_development_2014} & 0.35 & 40.0 & 0.01 & $40 \times 40 \times 40$ & \checkmark &  \\
Hiwonder & Jetauto \cite{roboworks_mecabot_pro_roboworks_2024} & 0.50 & 650.0 & 4.00 & $324 \times 659 \times 260$ &  & \checkmark \\
Hangfa & Compass Q2 \cite{hangfa_compass_q2_hangfa_2024} & 0.75 & 3,500.0 & 16.00 & $430 \times 330 \times 115$ &  & \checkmark \\
KUKA & youBot \cite{bischoff_kuka_2011} & 0.80 & 30,000.0 & 20.00 & $580 \times 380 \times 140$ &  & \checkmark \\
Agilex & Limo \cite{agilex_limo_agilex_2024} & 1.00 & 3,600.0 & 4.80 & $251 \times 220 \times 322$ &  & \checkmark \\
Clearpath & Ridgeback \cite{clearpath_robotics_inc_ridgeback_2020} & 1.10 & 50,000.0 & 135.00 & $790 \times 960 \times 310$ &  & \checkmark \\
Hangfa & Navigator Q2 \cite{hangfa_navigator_q2_hangfa_2024} & 1.20 & 20,300.0 & 35.00 & $520 \times 480 \times 240$ &  & \checkmark \\
Clearpath & Dingo-O \cite{clearpath_robotics_inc_dingo_2019} & 1.30 & 10,000.0 & 13.00 & $680 \times 510 \times 111$ &  & \checkmark \\
RoboWorks & Mecabot X \cite{roboworks_mecabot_x_roboworks_2024} & 1.32 & 6,000.0 & 20.50 & $630 \times 581 \times 203$ &  & \checkmark \\
Lynxmotion & A4WD3 \cite{lynxmotion_a4wd3_lynxmotion_2024} & 1.40 & 900.0 & 5.55 & $372 \times 374 \times 152$ &  & \checkmark \\
RDK & X3 \cite{rdk_x3_rdk_2024} & 1.50 & 590.0 & 1.93 & $181 \times 236 \times 185$ &  & \checkmark \\
Agilex & Ranger Mini \cite{agilex_ranger_mini_agilex_2024} & 1.50 & 12,400.0 & 64.50 & $500 \times 738 \times 338$ &  & \checkmark \\
RoboWorks & Mecabot Pro \cite{roboworks_mecabot_pro_roboworks_2024} & 1.83 & 2,600.0 & 10.80 & $541 \times 581 \times 225$ &  & \checkmark \\
Clearpath & Boxer \cite{clearpath_robotics_inc_boxer_2022} & 2.00 & 60,000.0 & 127.00 & $550 \times 750 \times 340$ &  & \checkmark \\
Festo & Robotino \cite{festo_robotino_festo_2024} & 2.70 & 8,000.0 & 22.80 & $450 \times 450 \times 325$ &  & \checkmark \\
Cambridge & \textbf{RoboMaster} (ours) & 4.45 & 689.0 & 3.00 & $240 \times 320 \times 200$ & \checkmark & \checkmark \\
\bottomrule
\end{tabular}

\end{table}

\autoref{fig:robot_platforms} visualises our survey in a 2D plot, and \autoref{tab:omnidir_robots} provides detailed statistics specifically for for omnidirectional drive robots.
We find that the majority of multi-agent research platforms utilize differential drive \cite{mondada_e-puck_2009,kernbach_re-embodiment_2009,arvin_mona_2019,paull_duckietown_2017,wilson_pheeno_2016,mclurkin_robot_2014,soares_khepera_2016,arvin_development_2009,caprari_design_2003}, thus positioning them in the bottom left corner of \autoref{fig:robot_platforms}. The reason is that such platforms tend to be designed to minimize cost, with tens of such agents in mind.
The slowest of these platforms is Alice~\cite{caprari_design_2003} at \SI{4}{cm/s}, and the fastest is the Khepera~IV~\cite{soares_khepera_2016} with \SI{0.93}{m/s}.
Three ground-based platforms are specifically geared towards multi-agent research but use dynamics other than differential drive, specifically the Minicar~\cite{hyldmar_fleet_2019},
Cellulo\cite{ozgur_cellulo_2017},
and Colias~\cite{arvin_development_2014}.
More recently, slightly faster commercial differential drive robots have become popular~\cite{clearpath_robotics_inc_jackal_2020,clearpath_robotics_inc_turtlebot_2022,turtlebot3_turtlebot3_2024,directdrivetech_diablo_2024,husarion_rosbot_husarion_2024}, ranging from \SI{0.22}{m/s} for the Turtlebot~3~\cite{turtlebot3_turtlebot3_2024} to \SI{2}{m/s} for the Diabolo~\cite{directdrivetech_diablo_2024}.

From the bottom left, we can see two clusters emerging towards the top and the right. The omnidirectional drive robots~\cite{ozgur_cellulo_2017,arvin_development_2014,roboworks_mecabot_pro_roboworks_2024,hangfa_compass_q2_hangfa_2024,bischoff_kuka_2011,clearpath_robotics_inc_ridgeback_2020,hangfa_navigator_q2_hangfa_2024,clearpath_robotics_inc_dingo_2019,roboworks_mecabot_x_roboworks_2024,rdk_x3_rdk_2024,agilex_ranger_mini_agilex_2024,roboworks_mecabot_pro_roboworks_2024,clearpath_robotics_inc_boxer_2022,festo_robotino_festo_2024,lynxmotion_a4wd3_lynxmotion_2024,agilex_limo_agilex_2024} tend to move faster than differential drive, ranging from \SI{0.15}{m/s} (Cellulo~\cite{ozgur_cellulo_2017}) to \SI{2.7}{m/s} (Festo Robotino~\cite{festo_robotino_festo_2024}), but are also significantly more expensive, with a mean cost of \usd{13k}.

Aerial platforms (such as multirotors) on the other hand are moderately priced at a mean of \usd{1600}, and can move at much higher speeds of up to \SI{36}{m/s} (Agilicious~\cite{foehn_agilicious_2022}), but come with other disadvantages compared to ground robots, such as operational overheads like short runtime (minutes as opposed to hours) or more challenging infrastructure needs (recovery and safety mechanisms and the availability of accurate indoor tracking).
These issues become particularly noticeable and more pronounced when developing multi-agent systems -- the feasibility of an experiment depends on all robots concurrently operating without faults and risks.
In a multi-robot system, individual robots requiring additional operator attention and overhead can thus pose a significant barrier to rapid deployment and testing.

To the right of the plot, we observe a cluster of Ackermann drive robots \cite{hyldmar_fleet_2019,balaji_deepracer_2020,karaman_project-based_2017,gonzales_autonomous_2016,agilex_titan_agilex_2024,liniger_optimization-based_2015,agilex_hunter_agilex_2024,williams_aggressive_2016}. These platforms move from top speeds ranging from \SI{0.7}{m/s} for the Minicar~\cite{hyldmar_fleet_2019} (\usd{76.5}) to \SI{8}{m/s} for the AutoRally~\cite{williams_aggressive_2016} (ca. \usd{10k}), with an average of \usd{4850}. While some of these platforms are agile, they tend to be expensive and not suitable for constrained indoor spaces.

Our solution sits at a frontier beyond which agility (speed) is attained only by aerial platforms, and a lower cost is only attained by sacrificing agility.
Furthermore, as we will describe in later sections, our software stack supports a wide range of relevant research problems, and has a proven track record throughout six previous publications over the last three years \cite{blumenkamp_framework_2022,bettini_heterogeneous_2023,blumenkamp_see_2023,gao_online_2023,yang_efficient_2023,blumenkamp_covis-net_2024}.

\section{Technical Details}
This section details the background and design decisions made when developing our robot platform.
We also cover all the supporting network, control and simulation infrastructure that complement it, and thus enable rapid experimentation.

\subsection{Platform}
We selected the DJI RoboMaster S1 as our platform's base, leveraging its design inspired by the RoboMaster competition—a DJI initiative to foster technological exchange and innovation among regional universities~\cite{committee_rmoc_robomaster_2022}. The S1, an educational robot, integrates a tank-like structure with a spring-dampened beam axle front suspension and a four-wheel omnidirectional drive system, and is designed for indoor use. It features a gimbal-mounted blaster toy gun and an onboard computer. DJI specifies the base platform's maximum speeds as \SI{3.5}{m/s} forward, \SI{2.5}{m/s} backward, \SI{2.8}{m/s} laterally, and \SI{600}{deg/s} rotationally, with a weight of 3.3 kg (note that as part of our contribution, we side-step these manufacturer-imposed restrictions). Each wheel has a diameter of \SI{100}{mm}, is powered by a brushless motor with integrated gears and a driver and supports closed-loop speed control up to 1000 RPM and a maximum torque of 0.25 Nm. This gives the platform a theoretical maximum speed of $(\pi \cdot \SI{0.1}{m} \cdot \SI{1000}{RPM}) / (\SI{60}{s/M}) = \SI{5.23}{m/s}$. The system is powered by a \SI{25.91}{Wh} smart battery, offering \SI{2.4}{Ah} capacity. \cite{dji_robomaster_s1_dji_2024}.

\begin{figure}[t]
    \centering
    \includegraphics[width=\linewidth]{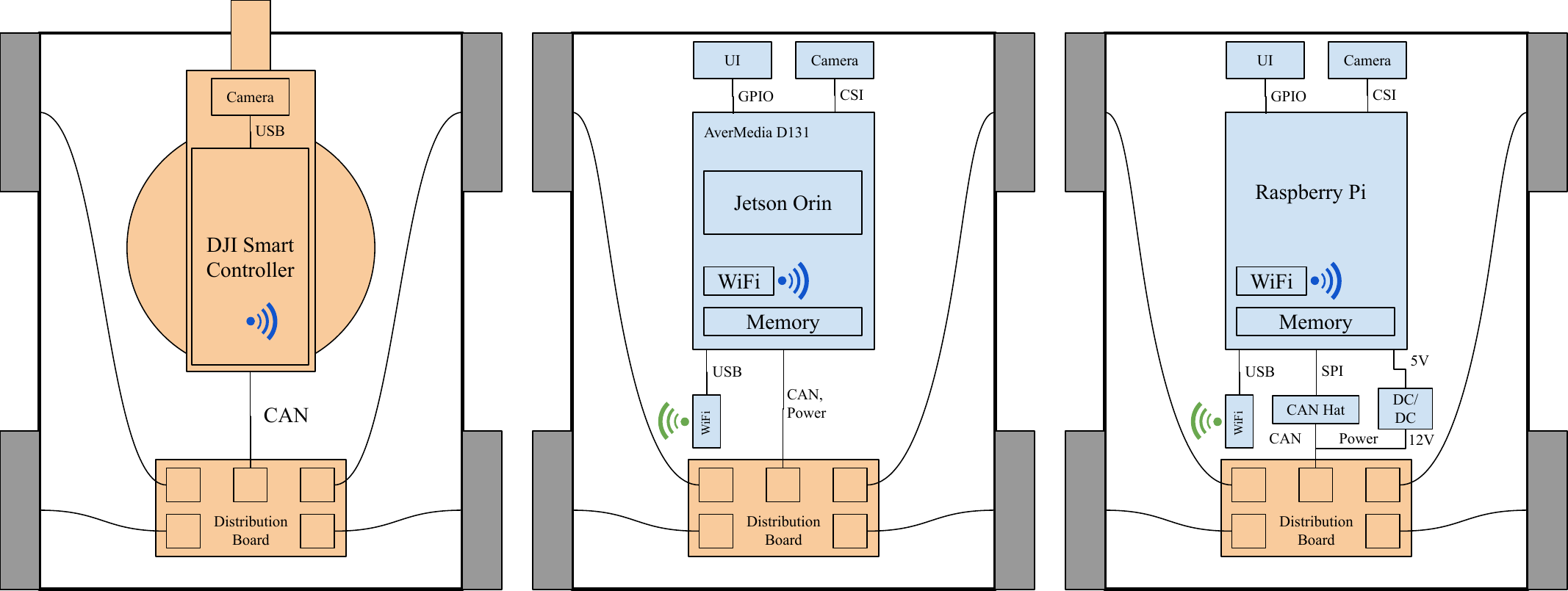}
    \caption{We modify the RoboMaster S1 Platform by removing the USB Camera, Smart Controller, Blaster and Turret and by replacing them with either a Jetson Orin on an AverMedia D131 carrier board or a Raspberry Pi. The platform is equipped with two WiFi modules, one for infrastructure communication (dark blue) and one specifically chosen for ad-hoc peer-to-peer communication (green).}
\label{fig:hardware}
\end{figure}

The onboard computer interfaces with a sub-controller, managing all sensors and wheels.
We replaced the onboard computer, gimbal, and blaster gun with a custom computing solution.
We laser-cut components from acrylic to mount our custom expansion in place of the turret. We provide a detailed overview of the compute configurations in \autoref{fig:hardware}.
All our design files are also part of our open-source release.
\footnote{\url{https://github.com/proroklab/cambridge-robomaster}}

\subsection{Compute}
The stock onboard compute module on the RoboMasters is impractical for running custom algorithms or implementing additional interfaces to cameras, sensors, networks, etc.
We therefore upgraded the onboard computer to a more powerful and versatile single board computer (SBC), focusing on the NVidia Jetson Orin family for its recency and lifecycle availability until January 2030\footnote{\url{https://developer.nvidia.com/embedded/lifecycle}, accessed 24/02/22}.
While the Jetson Orin NX 16GB was selected for its robust processing capabilities and suitability for our research, our design easily admits other cost-effective single-board computers such as the Raspberry Pi.
The Jetson Orin series necessitates a carrier board.
Our choice, the AverMedia D131 carrier board, was influenced by the availability of a CAN interface -- essential for communicating with the RoboMaster sub-controller -- and an optimal balance of interface availability and cost.
It accommodates two PCIe M.2 slots, where we installed a 256 GB SSD and an Intel 9260 WiFi module.
In an alternative Raspberry Pi setup, we used the built-in ports and interfaces and added an additional CAN shield for communication with the RoboMaster base. We provide a detailed bill of material in \autoref{tab:components}.

\begin{table}[t]
    \scriptsize
    \centering
    \caption{We provide a cost breakdown for six different compute and sensing configurations for our platform at the time of paper writing. Optional equipment, such as sensors, are put in parenthesis. All costs are stated in USD.}
    \label{tab:components}
    \setlength{\tabcolsep}{8pt}
    \begin{tabular}{l|ccc}
        \toprule
        Item & Raspberry Pi & Orin Nano & Orin NX \\
        \midrule
        Robot Base: RoboMaster S1 & 549 & 549 & 549\\
        Compute & 55 & 255 & 670 \\
        Carrier: AverMedia D131 & & 185 & 185 \\
        WiFi: Intel 9260 & & 15 & 15 \\
        Mem: Micron MTFDKBA256TFK & & 65 & 65 \\
        Mem: SD Card 256 GB & 30 & & \\
        Infrastructure: DC/DC Converter & 20 & & \\
        Infrastructure: Waveshare CAN Hat & 15 & & \\
        Infrastructure: Acrylics & 10 & 10 & 10 \\
        Infrastructure: Mounting & 10 & 10 & 10 \\
        Infrastructure: OLED Display & (10) & (10) & (10) \\
        WiFi: Netgear A6210 & (50) & (50) & (50) \\
        Sensing: Pi HQ Camera & (50) & (50) & (50) \\
        Sensing: Fisheye Lens & (20) & (20) & (20) \\
        \midrule
        Total Base & \textbf{689} & 1089 & 1504 \\
        Total Optional & (819) & (1219) & (1633) \\
        \bottomrule
    \end{tabular}
\end{table}

\subsection{CAN Communication}

\begin{figure}[t]
    \centering
    \includegraphics[width=\linewidth]{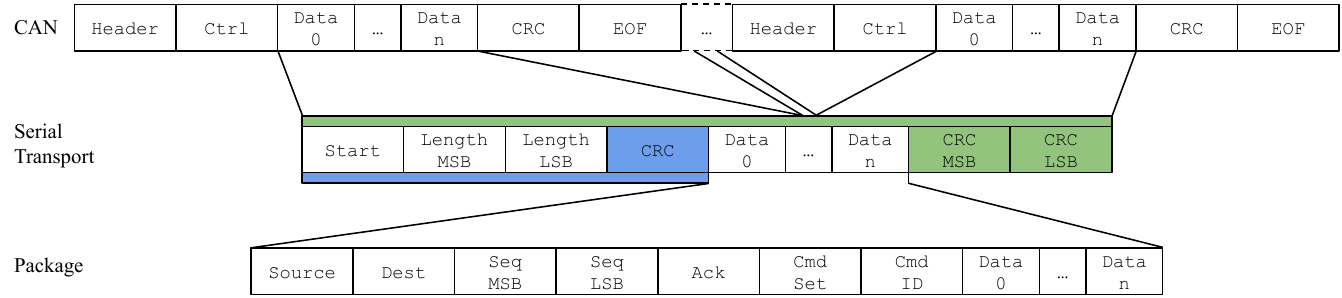}
    \caption{The CAN-based protocol used in the RoboMasters. DJI uses CAN as transport layer to allow multiple connected devices on the bus to broadcast information. Due to the limited length of CAN frames, a serial transport protocol is built on top, with a higher-level package definition permitting publisher/subscriber-type communication architectures.}
\label{fig:can_protocol}
\end{figure}

DJI offers an API for the RoboMaster EP, the educational counterpart to the RoboMaster S1, which enables control via a secondary computer (e.g., a Raspberry Pi) through USB.
However, this platform is not listed for sale on the DJI store, is unavailable in many parts of the world (including the UK), and at around \usd{1500}, is significantly costlier than the S1 model.
The API, which DJI has made open-source\footnote{\url{https://github.com/dji-sdk/RoboMaster-SDK}, accessed 03/19/24}, served as the foundation for our reverse engineering of the communication protocol between the main computer and the sub-controller.
DJI employs a Controller Area Network (CAN) for hardware layer transport but overlays it with a simplified serial package-based protocol.
This design likely addresses the limitations of CAN's 8-byte data size per packet while facilitating multi-node network connectivity, leveraging CAN's collision and arbitration management capabilities.
We outline the protocol's structure in \autoref{fig:can_protocol}, noting its use of a higher-level transport layer to specify node source and destination IDs, sequence counters for package tracking, various command sets and IDs, and acknowledgements, enabling subscription to specific package IDs for continuous data stream over the network. We release our C++ implementation of the messaging protocol as a stand-alone repository.\footnote{\url{https://github.com/proroklab/robomaster_sdk_can}}

\subsection{Infrastructure}

\begin{figure}[t]
    \centering
    \includegraphics[width=\linewidth]{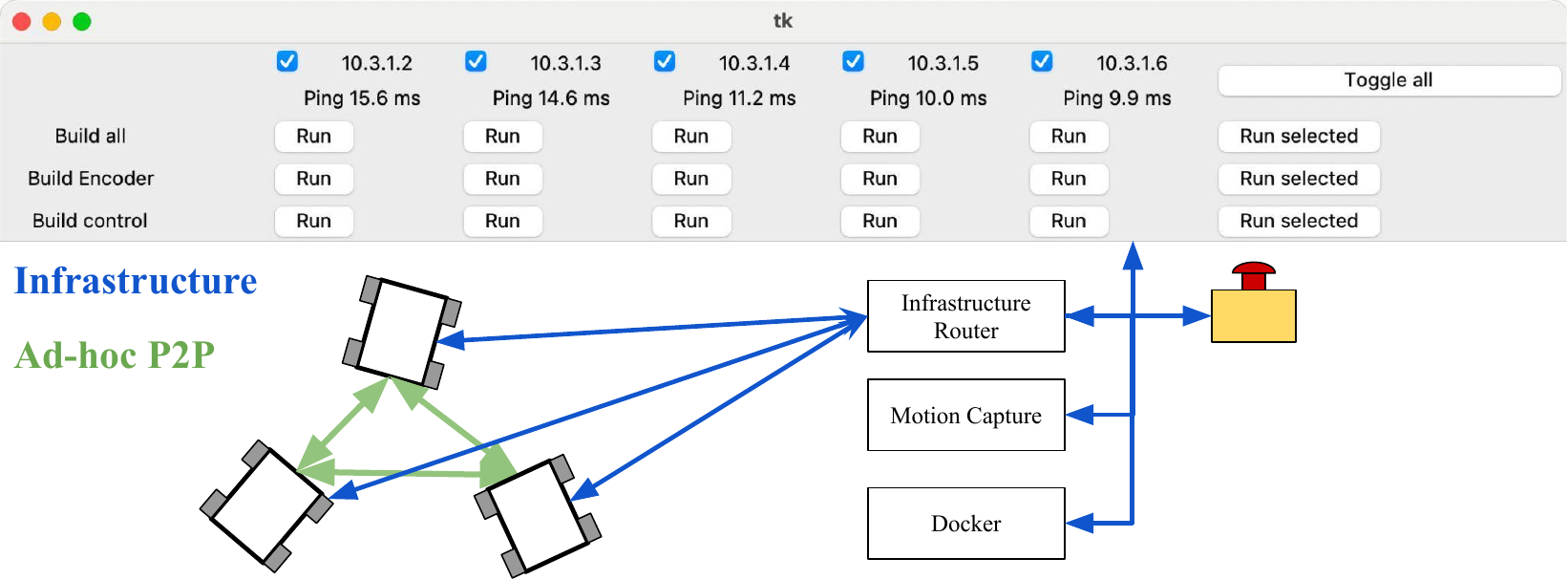}
    \caption{We utilize two wireless networks, one for infrastructure (blue) and one for ad-hoc peer-to-peer communication between agents. The infrastructure network is used to communicate with fixed infrastructure. The multi-agent user interface to control multiple agents efficiently is displayed in the top. The ad-hoc network is used for experiments requiring decentralized communication.}
\label{fig:infrastructure}
\end{figure}

Here we describe components of our supporting infrastructure, each of which is a product of over 4 years of refining and development.
The high reliability of these subsystems has been vital to our work in deploying multi-agent research. We provide a high-level overview of the infrastructure in \autoref{fig:infrastructure}

\subsubsection{Wireless Communication.} Our system employs dual network interfaces to facilitate distinct communication channels:
a backbone \textit{infrastructure} network for all participating systems, and, an \textit{ad-hoc} network exclusively for inter-robot communication.
For infrastructure-related communication, we utilize an Intel 9260 PCIe module for the Jetson and the integrated WiFi module for the Raspberry Pi that connects to a central WiFi router (described below).
This setup integrates with broader infrastructure components, such as desktop systems operating the Multi-Robot Manager UI, a motion capture system, and essential hardware like an emergency stop button.
This network also allows for SSH access into the robots, file sharing, and centralized ROS messaging.
In scenarios requiring only centralized control, a desktop machine executes our entire control stack (\textit{Freyja}, described later) for each robot, and transmits low-level wheel commands to each robot via ROS2.

For direct robot-to-robot communication, we use Netgear A6210 USB3.0 modules to establish an adhoc multicast network. This configuration does not rely on external infrastructure, enabling seamless peer-to-peer interactions.

\subsubsection{Network.}
The deployment of an enterprise-grade router proved crucial for centralized robot operation. We opted for a Mikrotik Router, selected for its superior configurability and flexible antenna setup. We employ a structured IP assignment strategy and bridge appropriate subnetworks that contain different categories of devices (robots, computers, guest devices, etc).

Our router and ad-hoc setups typically operate at a broadcast channel capacity of \SI{26}{Mb/s}, and we typically observe around \SI{1}{Mb/s} network load when operating 5 robots concurrently in a centralized mode (transmitting wheel velocity commands to each robot at 50 Hz). For details on our Ad-hoc peer-to-peer network, we refer the reader to~\cite{blumenkamp_framework_2022} and~\cite{blumenkamp_covis-net_2024}.

\subsubsection{Emergency Stop.}
Crash prevention is a priority in high-velocity multi-robot experiments, and can be particularly complex to implement in a fully decentralized framework.
Inevitable system latencies in re-establishing SSH sessions and the finite interval between terminating a control process and robot immobilization necessitate a more reliable stop mechanism.
Consequently, we integrate an emergency stop service into our RoboMaster ROS driver, accessible via the infrastructure network, that immediately halts velocity commands at the lowest level.
Additionally, we developed a stand-alone physical emergency stop button, powered by a Raspberry Pi Zero, that connects to the infrastructure network and features color LEDs to display the connectivity of the robots.

\subsubsection{On-Robot User Interface.} Incorporating on-robot displays has proven advantageous, especially for operations outside the conventional laboratory setting where an infrastructure network may be absent.
These displays, compact and economical at $128 \times 64$ pixels, present vital statistics including CPU and memory usage, CPU temperature, network details (such as IP addresses and interface connectivity), battery status, and the robot's hostname. An integrated button activates the emergency stop service if the robot is in motion. This feature is particularly useful in field deployments, offering a direct control and monitoring mechanism when network access is unavailable.

\subsubsection{ROS2 Drivers} We implemented a custom ROS2 Driver that accepts both high-level velocity commands in the robot's body frame as well as low-level wheel speed commands.
We use \texttt{systemd} to start these drivers automatically during the system boot to
make the experimental setup seamless.

\subsubsection{Decentralized Deployment} For deploying code across multiple robots, we utilize the infrastructure network, employing \texttt{rsync} over SSH to synchronize updates from a central workstation to each robot.
We extensively leverage Docker containers to remove experimental cross-interference among different users and setups.
This approach is particularly beneficial on the NVidia Jetson platform, for which an extensive library of pre-configured containers is available. These containers include environments for ROS, ROS2, PyTorch, and various other learning frameworks, either standalone or in combination\footnote{\url{https://github.com/dusty-nv/jetson-containers}, accessed on 20/03/2024.}. Docker is also configured to manage local image repositories within the infrastructure network, facilitating efficient image sharing among robots.

\subsubsection{Multi-Robot User Interface} We developed a configurable Python-based User Interface (UI) operated from a workstation linked to the infrastructure network. This UI facilitates connections to multiple nodes through SSH and supports scripting through a configuration file. It is designed to automate the parallel execution of repetitive commands that normally need to be executed in different terminals on different devices, thus significantly reducing the manual overhead associated with logging into individual robots, sourcing workspaces, and running launch files, especially when identical commands are required on multiple robots. A snapshot of the UI is shown in \autoref{fig:infrastructure}.

\subsection{Sensors}
Our platform is equipped with an array of exteroceptive and proprioceptive sensors, supported by ROS2 drivers for seamless integration.

\subsubsection{Proprioceptive} The DJI RoboMaster features an onboard suite of sensors including a 3D attitude sensor, wheel encoders, body velocity sensors, and battery level sensors, all accessible through the CAN interface.

\subsubsection{Exteroceptive} The D131 carrier board accommodates two camera CSI inputs. We utilize this capability by attaching a Raspberry Pi HQ camera, equipped with a fisheye lens, delivering a rectified image with a 120$^\circ$ field of view (FOV) using OpenCV. Additionally, the RoboMaster is outfitted with physical bumper sensors for enhanced interaction with its environment.

\subsection{Control: Freyja}
A necessary step in successful multi-robot deployments is the dependability of individuals to track their target trajectories accurately.
Towards this end, our onboard control stack, \textit{Freyja}~\cite{shankar_freyja_2021}, comprises of an optimal feedback controller and a library of state filters developed entirely in C++ over the ROS2 middleware.
The model-based controller is a linear quadratic regulator (LQR) that regulates the position and velocity of the robot along a trajectory by commanding wheel speeds for the low-level driver.
Denoting
$\boldsymbol{x} \equiv [\vec{p}, \vec{v}]^\top$ as the state variable (with $\vec{p}$ and $\vec{v}$ as the world-frame position and velocity vectors), we can write a dynamic model for the system as
$\dot{\boldsymbol{x}} = A\boldsymbol{x} + B\boldsymbol{u}$,
where $A$ and $B$ are system matrices, and $\boldsymbol{u}$ is the applied control input.
Then, for some reference state $\boldsymbol{x}_{\mathrm{ref}}$, it is possible to design a matrix $K$ for the control law, $\boldsymbol{u} = -K(\boldsymbol{x} - \boldsymbol{x}_{\mathrm{ref}})$ that stabilizes the system to that (possibly time-varying) reference state.

\subsection{Simulation: VMAS}
\label{sec:vmas}
Simulation is essential in multi-robot research as it enables safety and efficiency in data collection for machine learning pipelines. 
State-of-the-art simulators employ GPU vectorization to massively parallelize this process~\cite{mittal_orbit_2023}, leveraging the Single Instruction Multiple Data paradigm.
Towards this end, we provide a model for the RoboMaster platform in the Vectorized Multi-Agent Simulator (VMAS)~\cite{bettini_vmas_2022}. VMAS is a simulator consisting of a PyTorch differentiable physics engine and a set of challenging multi-robot tasks. Thanks to this, it is possible to rapidly train and deploy multi-robot policies from the simulator to the real world in a zero-shot manner, as demonstrated in previous work~\cite{bettini_heterogeneous_2023} (video \href{https://youtu.be/1tOIMgJf_VQ?si=iuIHak4syX4hNhaO&t=549}{here}).

To enable this deployment pipeline, we integrated several new features into the simulator. 
Firstly, to generate the $\vec{v}_{\mathrm{ref}}$ input required by Freyja, we implement a PID controller layer that regulates the input forces required by VMAS based on these reference velocities.
This layer allows us to train neural network policies that output desired velocities $\vec{v}_{\mathrm{ref}}$ and directly use them on our physical platform, without any change. Secondly, a key to the real-world deployment was an iterative fine-tuning process for some of the physics parameters in VMAS. These are parameters that depend on the robot platform and its interaction with the environment, such as friction coefficients, drag, control ranges, state boundaries, and robot geometry. This tuning process can be done once for each robot-environment pair and can be then utilized for any subsequent deployments.

A model for the RoboMaster platform in VMAS enables applications beyond prototyping in simulation and deploying in the real world. For instance, thanks to the accelerated computing available on our platform, policies trained in VMAS on approximated representations of the environment can be fine-tuned on the robot using data collected in the real world.

\section{Evaluations}

We now demonstrate the characteristics of our platform and the supporting suite of software framework through a wide range of experiments.
These explore the versatile and diverse nature of studies that this platform enables, such as,
\begin{itemize}[noitemsep,nolistsep]
  \item running classical model-based control (Freyja) for agile trajectory tracking,
  \item zero-shot sim-to-real transfer of multi-agent navigation tasks learnt in VMAS, 
  \item running distributed visual SLAM onboard to track relative position and thus avoid collisions, and, 
  \item utilizing large neural networks based on the DinoV2-s architecture to estimate relative poses and then track trajectories.
\end{itemize}

\noindent{}Finally, we summarize prior work from our research group that utilizes this platform for real-world deployments of multi-robot systems research.
\newline{}

\begin{figure}[t]
    \centering
    \includegraphics[width=\linewidth]{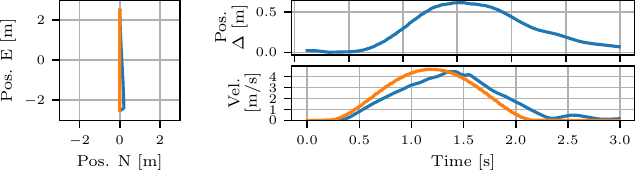}
    \includegraphics[width=\linewidth]{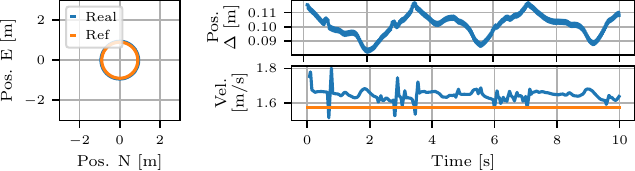}
    \caption{We demonstrate the efficiency of our platform combined with the Freyja controller suite in two single-agent trajectory tracking experiments, in a straight line trajectory (top) and a circular trajectory (bottom). We report the reference position and velocity (orange) and measured position and velocity (blue) as well as the tracking error dp over time. The position error increases and decreases since the reference acceleration is set to a higher value than the platform's physical limit.}
\label{fig:centr_trajectory}
\end{figure}

\noindent\textbf{Classical Trajectory Tracking.}
We first demonstrate the agility of our platform in a single-robot experiment in which we benchmark the tracking performance of Freyja as well as the velocity of our robot platform at high speeds. We demonstrate this on a straight line trajectory and a circle trajectory in \autoref{fig:centr_trajectory}. The straight line trajectory has a length of ca. \SI{5}{m}, and the circle trajectory has a diameter of \SI{1.5}{m}. We show that the robot reaches a top velocity of \SI{4.45}{m/s}, thus exceeding the maximum rated speed of the DJI RoboMaster platform by \SI{1.15}{m/s}, and accelerations of up to \SI{5}{m/s^2} in the straight line trajectory with a maximum error of \SI{0.5}{m}, and maintains a velocity of \SI{1.7}{m/s} on the circle trajectory with an average error of \SI{0.1}{m}. In the line trajectory, we are constrained by the size of the laboratory, which prohibits the robots from going faster.
The platform is capable of at least 600 deg/s in angular velocity, however, being a holonomic system, we do not stress this any further.
\newline{}

\begin{figure}[t]
    \centering
    \includegraphics[width=\linewidth]{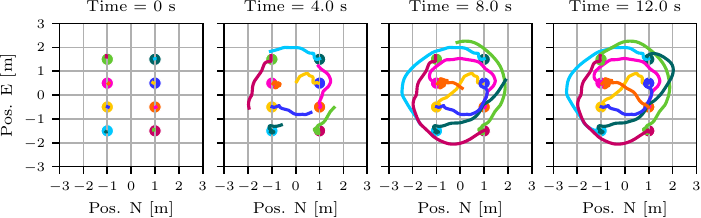}
    
    \hspace{0.9cm}
    \includegraphics[width=0.2\linewidth]{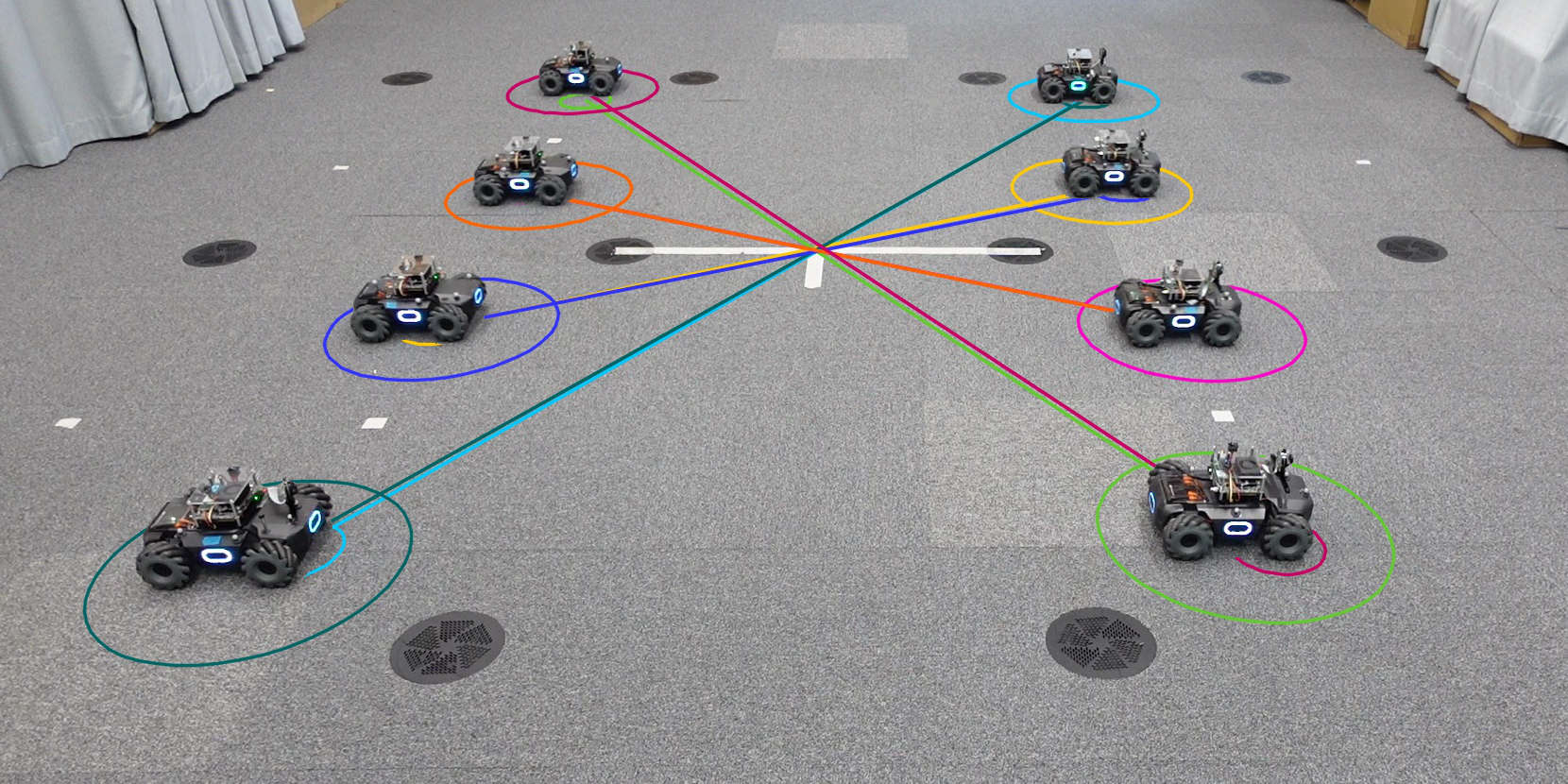}
    \hfill
    \includegraphics[width=0.2\linewidth]{figures/vmas_experiments/robomaster_line_b.jpg}
    \hfill
    \includegraphics[width=0.2\linewidth]{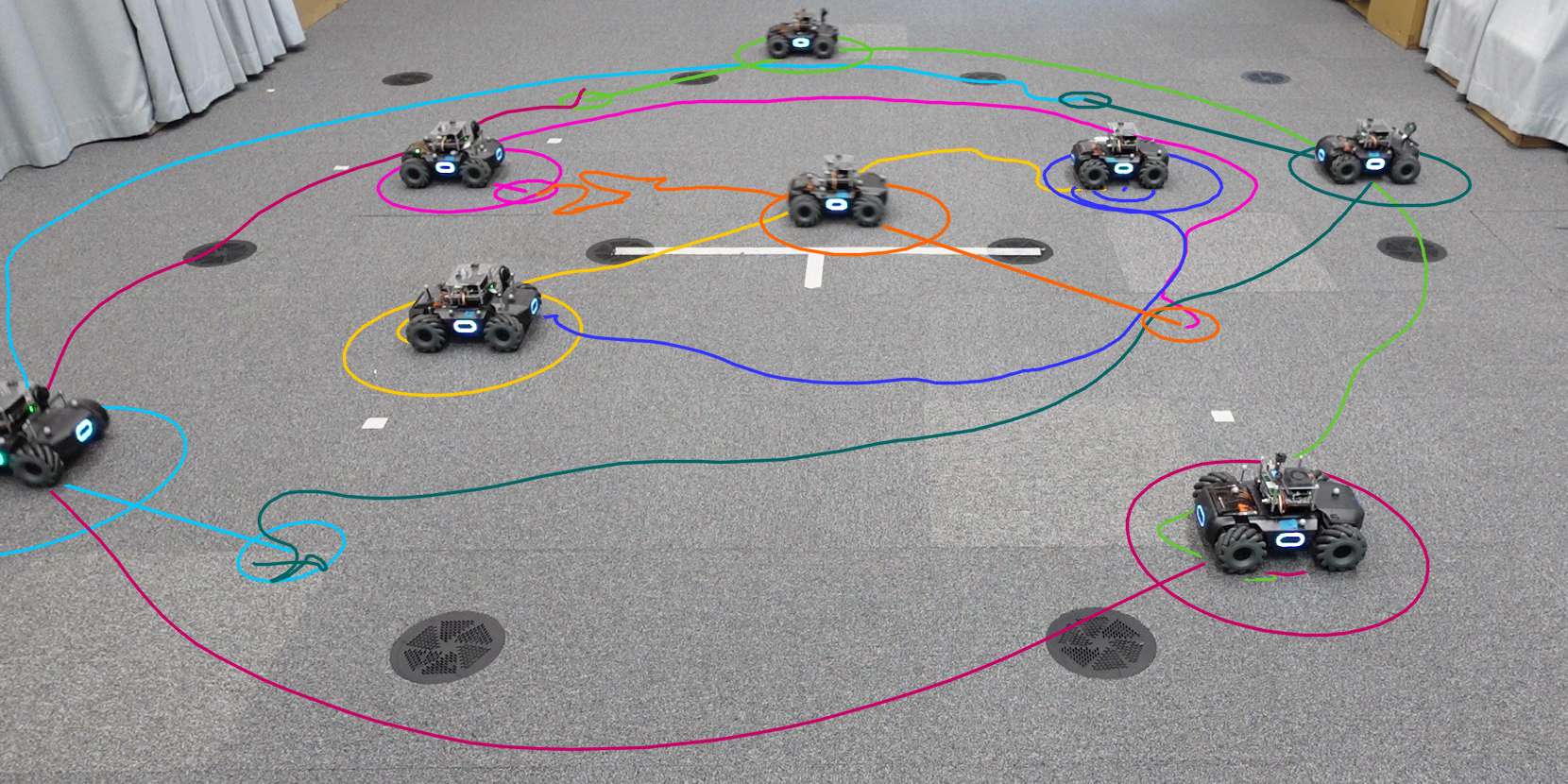}
    \hfill
    \includegraphics[width=0.2\linewidth]{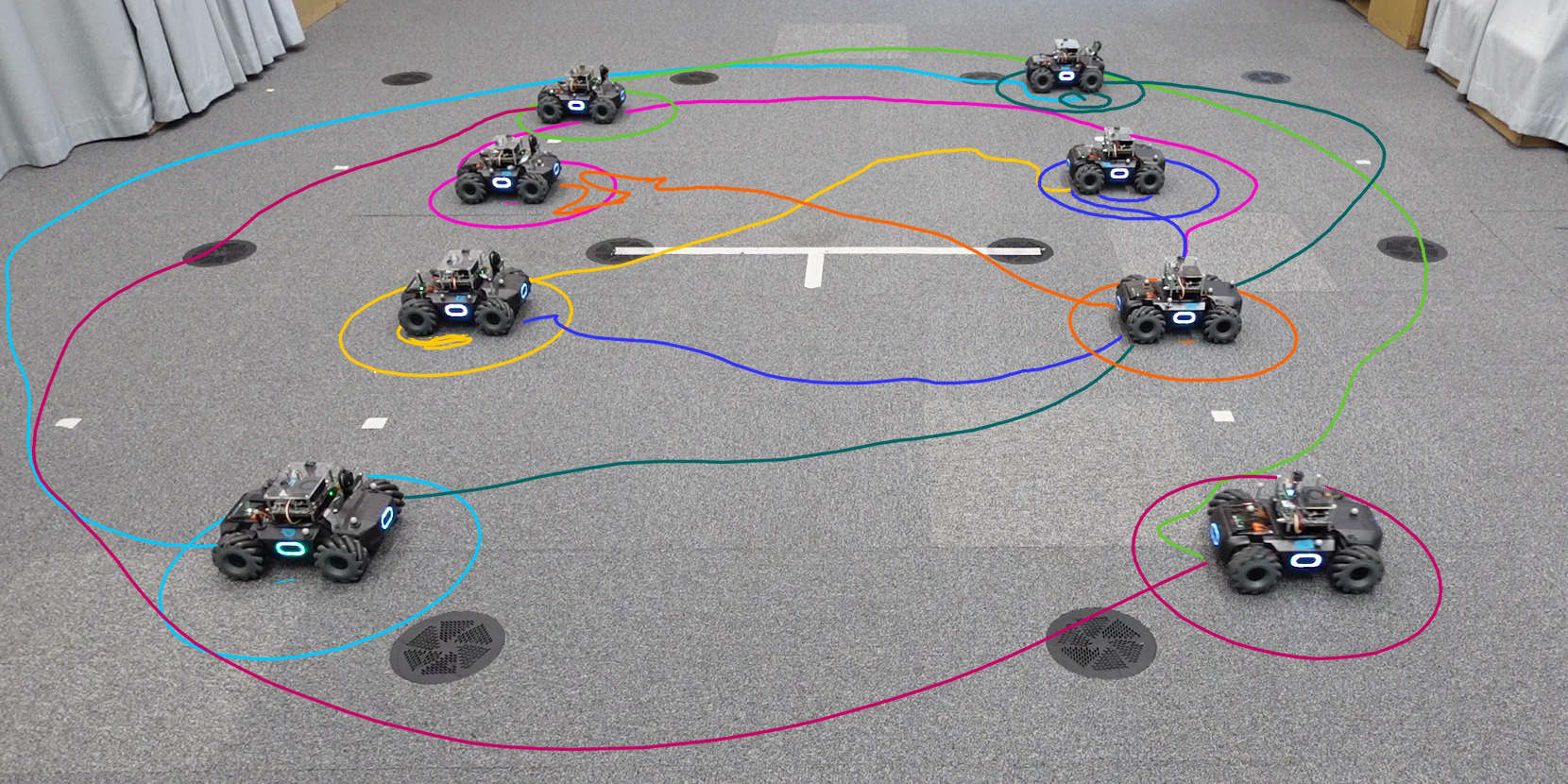}
    
    \caption{We showcase our sim-to-real capabilities on a MARL position swapping policy trained entirely in VMAS~\cite{bettini_vmas_2022}. Eight agents (color-coded) start on opposite sides of a \qtyproduct{4 x 2}{\metre} rectangle and attempt to swap positions with agents on the respective other side, which leads to multiple conflicts. The lines indicate their trajectories up to a particular time point, and the circles indicate their corresponding goals.
    We show snapshots from the real-world setup at the bottom.}
    \label{fig:vmas_swap}
\end{figure}

\noindent\textbf{Sim-to-real Multi-Agent Navigation.}
We now demonstrate zero-shot deployment of MARL policies
for a multi-robot navigation task 
trained in VMAS~\cite{bettini_vmas_2022} using the BenchMARL library~\cite{bettini_benchmarl_2023} and TorchRL~\cite{bou_torchrl_2024}, following the pipeline described in \autoref{sec:vmas}.
Agents are rewarded for navigating to their goal while avoiding collisions with each other, to do so they use the GPPO model (\cite{bettini_heterogeneous_2023}), which leverages a Graph Neural Network (GNN) in the policy for inter-agent communication. The GNN utilizes a 5-layer MLP with a total of 2300 trainable parameters, which can be evaluated on the Jetson Orin NX in \SI{0.5}{ms}. The results, reported in \autoref{fig:vmas_swap}, show that eight agents are able to seamlessly execute the learned policy collision-free in the previously unseen real-world environment.
\newline{}

\begin{figure}[t]
    \centering
    \includegraphics[width=\linewidth]{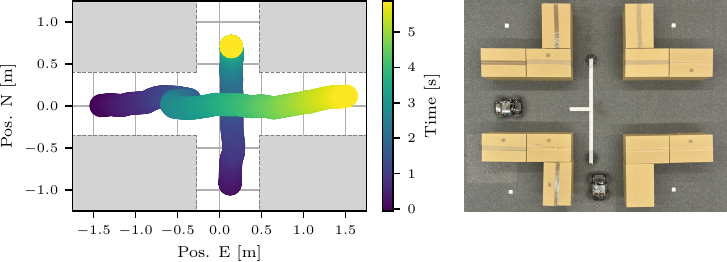}
    \caption{We demonstrate multi-agent collision avoidance, facilitated by a distributed visual SLAM system running locally on the robots. Two robots are set 90\textdegree{} to each other in an intersection environment (right). The agent travelling along the E axis is given a goal pose on the other side of the intersection, and successfully avoids a collision when the agent travelling along the N axis is pushed through the intersection. The trajectories generated by the SLAM system are presented on the left chart.}
\label{fig:collision_avoidance}
\end{figure}

\noindent\textbf{Distributed Visual SLAM.}
In this demonstration, we showcase the system's ability to run a decentralized visual SLAM system locally onboard using the optionally attached Raspberry Pi HQ cameras.
Our collaborative SLAM system enables the agents to share a \textit{unified} map of the world, thus facilitating relative positioning even when their views do not overlap, and even when the other agent is not explicitly detected in view.
To validate the quality of this shared map, we use
a nonlinear model predictive controller to avoid collisions with both peers and static obstacles.
\autoref{fig:collision_avoidance} tests a collision avoidance scenario at an intersection, where two robots (travelling along the horizontal and vertical axes) would nominally collide at the intersection.
However, they are able to localize each other close to the origin when common features are available in the environment, and the robot travelling along the vertical axis is able to slow down to avoid colliding. We are including additional details of this setup in a follow-up work.
\newline{}

\begin{figure}[tb]
    \centering
    \includegraphics[width=\linewidth]{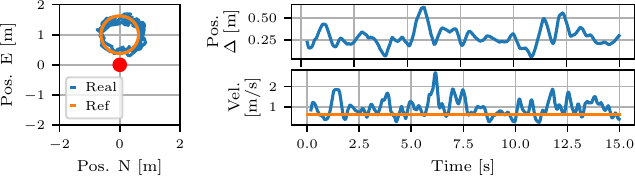}
    \caption{The decentralized trajectory tracking tracks a reference trajectory relative to a stationary robot (red dot), based on the relative poses estimated by a CoViS-Net~\cite{blumenkamp_covis-net_2024}. We run this neural network on board, in real-time. The position error and velocity are estimated from the predicted poses.
    }
\label{fig:rel_trajectory}
\end{figure}

\noindent\textbf{Multi-robot control using CoViS-Net.}
In this experiment, we deploy CoViS-Net~\cite{blumenkamp_covis-net_2024}, a pre-trained neural network to estimate relative poses between two camera frames, on two different robots. The camera images of two robots are encoded, and the features are broadcasted over the ad-hoc network to other robots within the communication range, based on which the relative pose can be predicted, which is then used to track a trajectory using Freyja.
In this setup, we use two robots, one that serves as origin for the reference coordinate frame of the other robot, which is commanded to move along a shifted circle trajectory.
The result, visualized in \autoref{fig:rel_trajectory}, shows that we are able to move with a velocity of about \SI{1}{m/s} on the circle, with an average estimated error of \SI{0.25}{m}.
Our model uses the DinoV2-s base neural network with 21M parameters, which we are able to run in \SI{20}{ms} on the Jetson Orin NX after optimising with NVidia TensorRT.\newline{}

\noindent\textbf{Case Studies.}
Our system has reached this level of maturity over the course of several iterations of research and development in the last four years.
Apart from the capabilities demonstrated above, this framework has featured as a primary research platform in several prior multi-agent works from our group:

\begin{itemize}[noitemsep,nolistsep]
    \item \textit{Multi-agent passage scenario}, where five RoboMasters navigate through a narrow passage, coordinated by onboard decentralized homogeneous GNN~\cite{blumenkamp_framework_2022};
    \item \textit{Real-world HetGPPO}, where heterogeneous GNNs trained in VMAS deployed on RoboMasters demonstrate superior resilience to real-world noise~\cite{bettini_heterogeneous_2023};
    \item \textit{Visual Navigation}, where a RoboMaster navigates an environment guided entirely using visual sensors and a policy trained in simulations~\cite{blumenkamp_see_2023};
    \item \textit{CBFs for multi-agent control}, where four RoboMasters use a control barrier function based GNN strategy to navigate~\cite{gao_online_2023}; and,
    \item \textit{Single-agent search \& navigation}, where LQR, CBF and RRT* are combined for safe and optimal single-robot motion planning~\cite{yang_efficient_2023}.
    \item \textit{Deployment of multi-agent foundation models}, where up to four RoboMaster estimate the relative pose to perform decentralized formation control in real-world scenarios \cite{blumenkamp_covis-net_2024}.
\end{itemize}
\section{Conclusion}
This article introduces a versatile omnidirectional ground robot system that is based on extensive modification and enhancement of the original DJI RoboMaster S1 platform.
We significantly extend the capabilities of the base platform by providing multiple options for more flexible and powerful compute solutions, onboard sensors, model-based control, and sim-to-real capabilities for distributed RL policies. 
After an intensive research and development pipeline over four years, our full platform (hardware and software) now serves as a highly reliable testbed well-suited for a wide range of experiments.
We showcase its capabilities through four new evaluations with fleet sizes of up to eight robots, and refer to six case studies from prior work that point to its proven track record.

\section*{Acknowledgments}
This work was supported by ARL DCIST CRA W911NF-17-2-0181 and European Research Council (ERC) Project 949949 (gAIa). J. Blumenkamp acknowledges the support of the ‘Studienstiftung des deutschen Volkes’ and an EPSRC tuition fee grant. We also acknowledge a gift from Arm. 
We further thank all other members of the Prorok Lab for their suggestions and helpful discussions that have contributed towards topics reported in this article.

\newpage
\bibliographystyle{plain}
\bibliography{author}

\end{document}